\documentclass[10pt]{article}
\usepackage{a4}
\usepackage{iwpt2001}
\newcommand{\newcite}[1]{\cite{#1}}
\author{%
	Detlef Prescher\\
  	DFKI Language Technology Lab\\
  	Stuhlsatzenhausweg 3, 66123 Saarbr\"ucken, Germany\\ 
  	{\tt prescher@dfki.de}
} 
\title{%
	INSIDE-OUTSIDE ESTIMATION MEETS DYNAMIC EM
}

\begin{document}

\newcommand{\qed}			
{\ensuremath{\mbox{ \textbf{q.e.d.}}}}

\newcommand{\states}			
{\ensuremath{\lbrace 1\ldots s\rbrace}}

\newcommand{\setOfStates}		
{\ensuremath{\mathcal{I}}}

\newcommand{\cats}			
{\ensuremath{\mathcal{C}}}

\newcommand{\clusgram}			
{\ensuremath{G_{\mbox{\tiny{cluster}}}}}

\newcommand{\clusprob}			
{\ensuremath{p_{\mbox{\tiny{cluster}}}}}

\newcommand{\base}			
{\ensuremath{\mbox{log-base }}}

\newcommand{\llh}			
{\ensuremath{L}}

\newcommand{\lhs}			
{\ensuremath{\mbox{lhs}}}

\newcommand{\rhs}			
{\ensuremath{\mbox{rhs}}}

\newcommand{\logicaland}		
{\ensuremath{\wedge\ }}

\newcommand{\logicalor}			
{\ensuremath{\vee\ }}

\newcommand{\argmax}[1]			
{\ensuremath{\mbox{argmax}_{#1}\ \ }}

\newcommand{\argmin}			
{\ensuremath{\mbox{argmin }}}

\newcommand{\cdata}			
{\ensuremath{{\cal{X}}}}

\newcommand{\cdat}			
{\ensuremath{x}}

\newcommand{\dist}			
{\ensuremath{p}}

\newcommand{\emp}			
{\ensuremath{\tilde{p}}} 

\newcommand{\entropie}			
{\ensuremath{H}}

\newcommand{\cross}			
{\ensuremath{H_{\mbox{\tiny{cross}}}}}

\newcommand{\estimator}			
{\ensuremath{\delta}}

\newcommand{\expect}[2]			
{\ensuremath{#1 \left[\ #2\ \right]}}

\newcommand{\kl}[2]			
{\ensuremath{D(#1 |\!| #2)}}

\newcommand{\events}			
{\ensuremath{{\cal{P}}(\cdata)}}

\newcommand{\event}			
{\ensuremath{X}}

\newcommand{\forest}			
{\ensuremath{{\mathcal{F}}orests}}

\newcommand{\inforest}			
{\ensuremath{{\mathcal{F}}_{\mbox{\tiny{inner}}}}}

\newcommand{\outforest}			
{\ensuremath{{\mathcal{F}}_{\mbox{\tiny{outer}}}}}

\newcommand{\freq}			
{\ensuremath{f}}

\newcommand{\freqs}			
{\ensuremath{{\cal{F}}}}

\newcommand{\idat}       		
{\ensuremath{y}} 

\newcommand{\idata}       		
{\ensuremath{{\cal{Y}}}}   

\newcommand{\mle}[1]			
{\ensuremath{\hat{#1}}} 

\newcommand{\emap}			
{\ensuremath{{\cal{M}}}} 

\newcommand{\emaux}[1]			
{\ensuremath{Q_{#1}}} 

\newcommand{\cond}[2]				
{\ensuremath{\dist_\oldparam({#1}|{#2})}}	

\newcommand{\nat}			
{\ensuremath{{\cal{N}}}}

\newcommand{\best}[1]			
{\ensuremath{#1\!\!-\!\!\mbox{best}}}  

\newcommand{\pair}[2]			
{\ensuremath{<\!\!#1,#2\!\!>}} 

\newcommand{\param}			
{\ensuremath{\Theta}}

\newcommand{\oldparam}			
{\ensuremath{\Phi}} 

\newcommand{\params}			
{\ensuremath{\Omega}}

\newcommand{\D}[2]			
{\ensuremath{\frac{\partial\ #1}{\partial\ #2}}}

\newcommand{\DD}[3]			
{\ensuremath{\frac{\partial^2 #1}{\partial #2 \partial #3}}}

\newcommand{\positives}       		
{\ensuremath{{\cal{R}}^+}}  

\newcommand{\powerset}       		
{\ensuremath{{\cal{P}}}}    

\renewcommand{\perp}       		
{\ensuremath{\mbox{perplexity}}}  

\newcommand{\dperp}       		
{\ensuremath{{\mbox{perplexity}}_D}}  

\newcommand{\hperp}       		
{\ensuremath{{\mbox{perplexity}}_H}}  

\newcommand{\real}       		
{\mathrm{I} \! \mathrm{R}}

\newcommand{\ringset}  		     	
{\ensuremath{{\cal{A}}}} 

\newcommand{\ringval}       		
{\ensuremath{{\mu_{\ringset}}}} 

\newcommand{\ringchart}       		
{\ensuremath{{chart_{\ringset}}}}    

\newcommand{\set}[2]			
{\ensuremath{\{#1,\ldots,#2\}}} 

\newcommand{\trees}			
{\ensuremath{{\mathcal{T}}}}

\newcommand{\treeroot}			
{\ensuremath{\mbox{root}}}

\newcommand{\treeyield}			
{\ensuremath{\mbox{yield}}}

\newcommand{\lng}			
{\ensuremath{{\mathcal{L}}}} 

\newcommand{\lexicon}			
{\ensuremath{\lbrace 1\ldots r \rbrace}} 

\newcommand{\totalFreq}			
{\ensuremath{|f|}}

\newcommand{\tuple}[2]			
{\ensuremath{<\!\!#1,\ldots,#2\!\!>}}  

\newcommand{\Frame}[1]			
{\fbox{\begin{minipage}{14.5cm}		
	#1				
\end{minipage}}}			

\newcommand{\theory}[2]			
{\ \newline
{{\fbox{{
\begin{minipage}{14.5cm}	
	\textbf{#1}:			
	#2				
\end{minipage}
}}}}
\newline}

\newcommand{\WithLines}[1]{
\
\newline
\begin{tabular}{l}
\hline
#1
\\
\hline
\end{tabular}
}

\newcommand{\algorithm}[3]		
{\Frame{				
\begin{description}			
\item[Input:] 			
	#1				
\item[Output:] 			
	#2				
\item[Procedure:]			
	\begin{enumerate}		
	#3				
	\end{enumerate}			
\end{description}			
}}	

\newcommand{\semiring}[6]		
{\Frame{				
\begin{description}			
\item[Elementmenge \ringset:]	
	#1				
\item[Addition $\oplus$:] 		
	#2				
\item[Multiplikation $\otimes$:]	
	#3				
\item[Additives neutrales Element 0:]	
	#4				
\item[Multiplikatives neutrales Element 1:]	
	#5				
\item[Werte gem\"{a}ss Grammatik:]  	
	#6				
\end{description}			
}}					

\newcommand{\cat}[1]			
{\mbox{\textit{#1}}}

\newcommand{\rewrite}
{\ensuremath{\rightarrow}} 

\newcommand{\with}
{, \ }  

\newcommand{\syn}
{\ensuremath{SY\!N}}     

\newcommand{\up}[1] 
{\ensuremath{ #1\!\uparrow }} 

\newcommand{\down}[1] 
{\ensuremath{ #1\!\downarrow }}  

\newcommand{\Math}[1]{
  \[
    #1
  \]
}

\newcommand{\Act}
{\ensuremath{p}}

\newcommand{\New}
{\ensuremath{\hat{p}}}  

\newcommand{\Reals}
{\mathrm{I} \! \mathrm{R}}

\newcommand{\RpowN}[1]
{{\mathrm{I} \! \mathrm{R}}^{#1}}

\newcommand{\Nats}
{\mathrm{I} \! \mathrm{N}}


\newcommand{\sep}			
{pa\-ra\-me\-ter\-se\-pa\-rier\-bar}

\newcommand{\EMA}			
{EM-Al\-go\-rith\-mus}

\newcommand{\EMK}[1]			
{EM\--ba\-siert#1 Klas\-si\-fi\-ka\-tions\-ver\-fahr\-en}

\newcommand{\EMM}[1]			
{EM\--ba\-siert#1 Klas\-si\-fi\-ka\-tions\-mo\-dell}

\newcommand{\SBA}			
{SB-Al\-go\-rith\-mus}

\newcommand{\GEM}			
{GEM-Al\-go\-rith\-mus}

\newcommand{\EMPA}			
{EM\--Pa\-ra\-me\-ter-\-Ab\-bild\-ung}

\newcommand{\EMAUX}			
{EM\--Hilfs\-funk\-tion}

\newcommand{\RVert}			
{Mar\-gi\-nal\-ver\-teil\-ung}

\newcommand{\RFreq}			
{Mar\-gi\-nal\-h\"auf\-ig\-keit}

\newcommand{\symbolisch}		
{sym\-bo\-lisch}

\newcommand{\AK}			
{Ana\-ly\-se\--Kom\-po\-nen\-te}

\newcommand{\hmm}			
{Hid\-den\--Mar\-kov\--Mo\-dell}

\newcommand{\cfg}			
{kon\-text\-freie Gram\-matik}

\newcommand{\prob}			
{pro\-ba\-bi\-lis\-tische}

\newcommand{\GR}			
{Gram\-matik\-re\-gel}

\newcommand{\GRW}			
{Gram\-matik\-re\-gel\--\Wahr}

\newcommand{\GRGEN}			
{Gram\-matik\-re\-gel\--Gen\-er\-ier\-ungs\-funk\-tion}

\newcommand{\TV}			
{Trai\-nings\-ver\-fah\-ren}

\newcommand{\BTV}			
{Baum\-bank\--\TV}

\newcommand{\Cross}			
{Cross\--En\-tro\-pie}

\newcommand{\LL}			
{Log\--Like\-li\-hood}

\newcommand{\loglin}			
{log\--li\-ne\-ar}

\newcommand{\komplett}			
{voll\-st\"{a}n\-dig}

\newcommand{\inkomplett}		
{un\-voll\-st\"{a}n\-dig}

\newcommand{\EmpWahr}			
{em\-pi\-ri\-sche \Wahr}

\newcommand{\empirische}		
{em\-pi\-ri\-sche}

\newcommand{\EmpVert}			
{em\-pi\-ri\-sche \Vert}

\newcommand{\Erw}			
{Er\-war\-tungs\-wert}

\newcommand{\Est}			
{Esti\-mier\-ung}

\newcommand{\EstVerfahren}		
{Es\-ti\-mier\-ungs\-ver\-fahren}

\newcommand{\Freq}			
{H\"{a}u\-fig\-keit}

\newcommand{\FreqVert}			
{\Freq s\-ver\-teil\-ung}

\newcommand{\Korp}			
{Kor\-pus}

\newcommand{\Entropie}			
{En\-tro\-pie}

\newcommand{\IO}			
{In\-side\--Out\-side}

\newcommand{\IOA}			
{\IO\--Algo\-rith\-mus}

\newcommand{\KorpPos}			
{\Korp\-position}

\newcommand{\KorpWahr}			
{\Korp\-wahr\-schein\-lich\-keit}

\newcommand{\RW}			
{Re\-gel\-wahr\-schein\-lich\-keit}

\newcommand{\KorpPerp}			
{\Korp\-per\-plex\-it\"{a}t}

\newcommand{\Perp}			
{\Korp\-per\-plex\-it\"{a}t}

\newcommand{\KorpVert}			
{\FreqVert\ aller \Typ en im \Korp}

\newcommand{\KL}			
{Kull\-back\--Leib\-ler\--Dis\-tanz}

\newcommand{\MLE}			
{Max\-imum\--Likeli\-hood\--\Est}

\newcommand{\MAPE}			
{Max\-imum\--A\--Post\-eriori\--\Est}

\newcommand{\Semiring}			
{Se\-mi\-ring}

\newcommand{\SPA}			
{Semi\-ring\--Par\-sing\--Algo\-rith\-mus}

\newcommand{\Token}			
{Da\-ten\-to\-ken}

\newcommand{\totale}			
{ab\-so\-lute}

\newcommand{\TotalFreq}			
{ab\-so\-lute \Freq}

\newcommand{\Typ}			
{Da\-ten\-typ}

\renewcommand{\Vert}			
{\Wahr s\-ver\-teil\-ung}
 
\newcommand{\Wahr}			
{Wahr\-schein\-lich\-keit}

\newcommand{\datum}			
{\mbox{
{\psfig{file=_figures_/datum-symbol.ps,angle=270,width=.3cm,height=.3cm}}
}}

\newcommand{\algo}			
{\mbox{
{\psfig{file=_figures_/algorithm-symbol.ps,angle=270,width=.6cm,height=.3cm}}
}}

\newcommand{\zustand}			
{\mbox{
{\psfig{file=_figures_/zustand.ps,angle=270,width=.6cm,height=.6cm}}
}}

\newcommand{\startzustand}			
{\mbox{
{\psfig{file=_figures_/startzustand.ps,angle=270,width=1.4cm,height=.6cm}}
}}

\newcommand{\transitzustand}			
{\mbox{
{\psfig{file=_figures_/transitstate.ps,angle=270,width=2.0cm,height=.6cm}}
}}

\newcommand{\emission}			
{\mbox{
{\psfig{file=_figures_/emission.ps,angle=270,width=2.2cm,height=.6cm}}
}}
\maketitle
\begin{abstract}
We briefly review the inside-outside and EM
algorithm for probabilistic context-free grammars. 
As a result, we formally prove that inside-outside estimation 
is a dynamic-programming variant of EM. 
This is interesting in its own right, but even more when
considered in a theoretical context since
the well-known convergence behavior of
inside-outside estimation has been confirmed by many experiments
but apparently has never been formally proved. 
However, being a version of EM, inside-outside
estimation also inherits the good convergence behavior
of EM. Therefore, the as
yet imperfect line of argumentation can be transformed into a coherent
proof.
\end{abstract}

\section{Inside-Outside Estimation}
\label{SecIO}

The modern \textbf{inside-outside algorithm} was introduced by 
\cite{Lari:90} who reviewed an algorithm proposed by
\cite{Baker:79} and extended it to an iterative training method for
probabilistic context-free grammars enabling the use of unrestricted 
free text. In the following, $y_1 \ldots y_N$ are numbered (but unannotated) 
sentences. 

\textbf{Definition:}
Inside-outside re-estimation formulas
for probabilistic context-free grammars in Chomsky normal form 
are given by (see \cite{Lari:90}, but see also \cite{Baker:79} for the
special case $N=1$):
\[
	\hat{p}(A \rightarrow a)
	:= 
	\frac
	{
		\sum_{w=y_1}^{y_N} 
		C_w(A \rightarrow a) 
	}
	{
		\sum_{w=y_1}^{y_N} 
		C_w(A) 
	}
	,
	\mbox{and }
	\hat{p}(A \rightarrow B C)
	:=
	\frac
	{
		\sum_{w=y_1}^{y_N} 
		C_w(A \rightarrow B C) 
	}
	{
		\sum_{w=y_1}^{y_N}
		C_w(A) 
	}
	.\
\]
The key variables of this definition are so-called
\textbf{category} and \textbf{rule counts}:
\(	
	C_w(A)
	:=
	{
		\frac{1}{P}
		\sum_{s=1}^n
		\sum_{t=s}^n
		e(s,t,A) \cdot f(s,t,A)
	}
	,\ 
\)
\(
	C_w(A \rightarrow a)
	:=
	\frac{1}{P}
	\sum_{1 \le t \le n,\ w_t = a}
		e(t,t,A) \cdot f(t,t,A)
	,\ 
\)
and 
\(
	C_w(A \rightarrow B C)
	:=
	\frac{1}{P}
	\sum_{s=1}^{n-1}
	\sum_{t=s+1}^{n}
	\sum_{r=s}^{t-1}
	p(A \rightarrow B C)
	e(s,r,B)
	e(r+1,t,C)
	f(s,t,A)
\)
which are computed for each sentence $w := w_1\ldots w_n$
with so-called inside and outside probabilities: 
An \textbf{inside probability} is defined as the probability of
category $A$ generating observations
$w_s\ldots w_t$, i.e.
\(
	e(s,t,A)
	:= 
	\dist(A \Rightarrow^* w_s \ldots w_t)
	.\ 
\)
In determining a recursive procedure for calculating $e$, two cases
must be considered:
\vspace{-0.5em}
\begin{itemize}
\item
	$(s=t)$: Only one observation is emitted and therefore a 
	rule of the form $A \rightarrow w_s$ applies:
\(
	e(s,s,A)
	=
	\dist(A \rightarrow w_s)
	,\ 
\)
if
\(
	(A \rightarrow w_s) \in G
\)
(and 0, otherwise). 
\vspace{-0.5em}
\item
	$(s < t)$: In this case we know that rules of the form 
	$A \rightarrow B C$ must apply since more than one observation
	is involved. Thus, $e(s,t,A)$ can be expressed as follows:
\(
	e(s,t,A)
	=
	\sum_{(A \rightarrow B C) \in G\ }
	\sum_{r=s}^{t-1}
	\dist \left( A \rightarrow B C \right)
	\cdot 
	e(s,r,B)	
	\cdot	
	e(r+1,t,C)
	.
\)	
\end{itemize}
The quantity $e$ can therefore be computed recursively by determining
$e$ for all sequences of length 1, then 2,
and so on. The sentence
probability $P := p(S \Rightarrow^* w)$ is 
a special inside probability. 
The \textbf{outside probabilities} are defined as follows:
\(
	f(s,t,A)
	= 
	\dist 
	\left( 
		S \Rightarrow^* w_1\ldots w_{s-1} A w_{t+1}\ldots w_n
	\right)
	.\
\) 
The quantity $f(s,t,A)$ 
may be thought of as the probability that  $A$ is generated in the re-write
process and that the strings not dominated by it are 
$w_1\ldots w_{s-1}$ to the left and $w_{t+1}\ldots w_n$ to the right.
In this case, the non-terminal $A$ could be one of two possible
settings
$C \rightarrow B\ A$  or $C \rightarrow A\ B$, hence: 
\(
	f(s,t,A)
	=
	\sum_{B,\ C \in G}
	\left(\ 
		\sum_{r=1}^{s-1}
		f(r,t,C)
		\cdot
		\dist(C \rightarrow B A)
		\cdot
		e(r,s-1,B)
	\right.
	+
	\left.
		\sum_{r=t+1}^{n}
		f(s,r,C)
		\cdot
		\dist(C \rightarrow A B)
		\cdot
		e(t+1,r,B)
	\right)
\)
and
\(
	f(s,t,A)
	=
	\left \lbrace
	\begin{array}{ll}
		1 
	&
		\mbox{ if } A = S
	\\
		0
	&
		\mbox{ else }
	\end{array}
	\right.
	.\ 
\)
After the inside probabilities have been computed bottom-up,
the outside probabilities can therefore be computed top-down.
Unfortunately, no convergence proofs of inside-outside estimation were 
given by \cite{Baker:79} 
and \cite{Lari:90}.

\section{EM for Probabilistic Context-Free Grammars}
\label{SecEM}

The EM algorithm was introduced by \cite{Dempster:77} as 
iterative maximum likelihood estimation for parameterized probability
models $p(y)$
using a sample $\emp(y)$ of 
\textbf{incomplete data types} $y$ which are defined via  
a \textbf{symbolic analyzer} $\event(y)$ 
dealing with \textbf{complete data types} $x$. 
It is known, that EM generalizes 
ordinary maximum likelihood estimation and  
monotonically increases
the log-likelihood
\(
	L(\dist) 
	:= 
	\sum_{y} 
	\emp(y) \cdot
	\log 
	\sum_{x \in \event(y)} 
	\dist(x)
	.
\)  
Furthermore, the limit point of a convergent parameter sequence 
is a stationary point (i.e. local minimum, saddle point or maximum) 
of the log likelihood \cite{Dempster:77}. Moreover, 
both the parameter sequence
and the associated sequence of log likelihood values converge (in some
cases to local maxima), 
if some weak conditions are fulfilled
\cite{Wu:83}.

Applying EM to probabilistic context-free grammars, the 
\textbf{grammatical sentences} $y$ are viewed as incomplete and 
their \textbf{syntax trees} $x$ as complete. The required 
symbolic analyzer is given by a \textbf{parser} 
computing all trees 
$x \in \trees(y)$ for a sentence $y$. 
Via these non-probabilistic EM components, 
the probability model for the sentences is defined as
\(
	p(y) :=
	\sum_{x \in \trees(y)} p(x)
	:=
	\sum_{x \in \trees(y)} \prod_{r} p(r)^{f_r(x)}
	,
\)
where $f_r(x)$ is the frequency of rule $r$ occuring in
$x$, and parameterization is 
given by \textbf{rule probabilties} $p(r)$.
The key variables of EM re-estimation are \textbf{conditional
expected frequencies} (relying on the conditional probability
$p(x|y) := \frac{p(x)}{p(y)}$)
for rules $r$ and categories $A$:
\(
	\expect{p(.|y)}{f_r} := 
	\sum_{x \in \trees(y)}
	\dist(x|y) \cdot f_r(x)
\)
and
\(
	\expect{p(.|y)}{f_A}
	:= 
	\sum_{x \in \trees(y)}
	\dist(x|y) \cdot f_A(x)
	,
\)
where $f_A(x) := \sum_{r \in G_A} f_r(x)$ is the frequency of  
category $A$ occuring in $x$, and $G_A$ is the set of grammar rules
with left-hand side $A$. 
See e.g. \cite{Prescher:DynamicEM}:

\textbf{Lemma:}
EM re-estimation formulas for probabilistic context-free grammars
are given by:
\[
	\mle{\dist}(r)
	=
	\frac
	{
		\expect{\emp}{\expect{p(.|.)}{f_r}}
	}
	{	
		\expect{\emp}{\expect{p(.|.)}{f_A}}
	}
	=
	\frac
	{
		\sum_{y} 
		\emp(y)
		\cdot
		\expect{p(.|y)}{f_r}
	}
	{	
		\sum_{y} 
		\emp(y)
		\cdot
		\expect{p(.|y)}{f_A}
	}
	\qquad 
	(r \in G,\ A=\lhs(r))
	\ .
\]

\section{Inside-Outside as Dynamic EM}
\label{SecDynamicEM}

In this section, the well-known convergence properties
of the inside-outside algorithm, which 
have been unfortunately omitted in the original 
literature (\newcite{Baker:79},
\newcite{Lari:90}),
will be formally proven. 
For this purpose, we will show that the inside-outside algorithm is
a dynamic-programming variant of the EM algorithm for context-free
grammars. This property is also well-known in stochastic linguistics, but
to the best of our knowledege all mentioned properties
have not been
formally proven till now.

\textbf{Theorem:}{
For a context-free grammar in Chomsky normal form,
let
$\mle{p}(r)$ 
be re-estimated rule probabilities resulting from one single step of the
inside-outside algorithm using the current rule probabilities
$p(r)$. Then:
(i) 
The log likelihood $\llh(.)$ of the training corpus 
increases monotonically, i.e.
\(
	\llh(\mle{\dist}) \ge \llh(\dist)
	.
\) 
(ii) 
The limit points of a sequence of re-estimated probabilities
are stationary points (i.e. maxima, minima or saddle points) 
of the log likelihood function.
(iii) The inside-outside algorithm is a dynamic-programming variant 
of the EM algorithm, i.e. $\mle{p}(r)$ corresponds to
$\mle{\dist}_{EM}(r)$ resulting from one
single EM iteration (using also $p(r)$ as current rule probabilities).
}

\textbf{Proof:} (i) and (ii) follow using both (iii) and the
convergence properties
of EM. (iii):
The empirical distribution of the sentences is defined as 
$\emp(y) = \frac{f(y)}{N}$, where $f(y)$ is the frequency of $y$
occuring in the corpus $y_1 \ldots y_N$. Thus,
for each rule $r$
with left-hand side $A$:
\(
	\mle{\dist}_{EM}(r)
	=
	\frac
	{
		\sum_{y=y_1}^{y_N} 
		\sum_{x \in \trees(y)}
		\dist(x|y) \cdot f_r(x)
	}
	{	
		\sum_{y=y_1}^{y_N}  
		\sum_{x \in \trees(y)}
		\dist(x|y) \cdot f_A(x)
	}
	\ .
\)
Comparing these formulas with the re-estimation formulas 
presented by \newcite{Lari:90},
it follows
\(
	\mle{\dist}_{EM}(r) \ = \ \mle{\dist}(r)
	,
\)
if for each sentence $y$,
for each rule $r$ and each category $A$
the following propositions can be shown:
\[
	C_y(r) =  \sum_{x \in \trees(y)} p(x|y) \cdot f_r(x),
	\mbox{and }\quad
	C_y(A) =  \sum_{x \in \trees(y)} p(x|y) \cdot f_A(x)\ . 
\]
This is the goal of the rest of the proof, which we split in two lemmas.
The first lemma is probably due to \cite{Charniak:93}, where  
corresponding formulas are used, but not explicitly proven,
to present inside-outside estimation.
The lemma says that category counts can be computed by summing
certain rule counts.

\textbf{Lemma: }{
\(
	C_y(A)
	=
	\sum_{r \in G_A}
	C_y(r)
\)
for each sentence $y$ and each category
$A$. 
}

\textbf{Proof:} Assuming Chomsky normal form, and $y=w_1\ldots w_n$: 
\begin{eqnarray*}
	\sum_{r \in G_A} 
	C_y(r)
	&=&
	\sum_{a} 
	C_y(A \rightarrow a)
	\ + \
	\sum_{B,C \in G} 
	C_y(A \rightarrow B \ C)
\\
	&=&
	\sum_{a} 
	\frac{1}{P}
	\sum_{1 \le t \le n,\ w_t = a}
		e(t,t,A) \ f(t,t,A)
\\
	&&+
	\sum_{B,C \in G} 
	\frac{1}{P}
	\sum_{s=1}^{n-1}
	\sum_{t=s+1}^{n}
	\sum_{r=s}^{t-1}
	p(A \rightarrow B C)
	e(s,r,B)
	e(r+1,t,C)
	f(s,t,A)
\\
	&=&
	\frac{1}{P}
	\left(
	\sum_{1 \le t \le n}
	e(t,t,A) \ f(t,t,A)
	\right.
\\
	&&+
	\left.
	\sum_{s=1}^{n-1}
	\sum_{t=s+1}^{n}
	f(s,t,A)	
	\sum_{B,C \in G} 
	\sum_{r=s}^{t-1}
	p(A \rightarrow B C)
	e(s,r,B)
	e(r+1,t,C)
	\right)
\\
	&=&
	\frac{1}{P}
	\left(
	\sum_{1 \le t \le n}
	e(t,t,A) \ f(t,t,A)
	\ + \
	\sum_{s=1}^{n-1}
	\sum_{t=s+1}^{n}
	f(s,t,A) \	
	e(s,t,A)
	\right)
\\
	&=&
	\frac{1}{P}
	\sum_{1 \le s \le t \le n}
	e(s,t,A) \
	f(s,t,A)
	\ = \
	C_y(A)
	\ .
\end{eqnarray*}
In the fourth equation, we used the recursion formula of the inside
probabilities. \qed

It follows
that the desired 
identities for the category counts can be calculated (by summation
over all rules with the same left-hand side) using the identities for 
the rule counts, since
\(
	C_y(A) 
	= 
	\sum_{A \rightarrow \alpha} 
	C_y( A \rightarrow \alpha)
	,
\)
and per definition
\(
	f_A(x)
	=
	\sum_{A \rightarrow \alpha} 
	f_{A \rightarrow \alpha}(x)
	\ .
\) 
Thus, the proof of the theorem is completed, 
as once as the following central lemma has been
proven.
It states that the counts of the inside-outside algorithm
can be identified 
with the expected rule frequencies of the EM algorithm.

\textbf{Lemma: }{
For each sentence $y$
and each rule $r$:\ \ 
\(
	C_y(r) 
	\ = \
	\sum_{x \in \trees(y)} p(x|y) \cdot f_r(x)	
	\ = \ 
	\expect{p(.|y)}{f_r}
	\ .
\)
}

\textbf{Proof:}
The second equation is simply the definition of the expectation.
Assuming Chomsky normal form, 
two cases must be considered.
First, the rule has the form 
$A \rightarrow B\ C$:

For a given sentence $y=w_1\ldots w_n$ and 
given three \textbf{spans} $(s,r,B)$, $(r+1,t,C)$, 
$(s,t,A)$ with
$1\le s \le r < t \le n$, let
$\event_{(s,t,A)(s,r,B)(r+1,t,C)}$ be the \textbf{parse forest}
corresponding to the following \textbf{derivation}: 
\(
	S 
	\Rightarrow^* w_1\ldots w_{s-1}\ A\ w_{t+1}\ldots w_n
	\Rightarrow   w_1\ldots w_{s-1}\ B\ C\ w_{t+1}\ldots w_n
	\Rightarrow^*   w_1\ldots w_{r}\ C\ w_{t+1}\ldots w_n
	\Rightarrow^*   w_1\ldots w_n
	.
\)
Let
\(
	f_{(s,t,A)(s,r,B)(r+1,t,C)}(x)
	\ := \
	\left\lbrace
	\begin{array}{cl}
		1 & \mbox{ if } \quad x \in \event_{(s,t,A)(s,r,B)(r+1,t,C)} 
	\\
		0 & \mbox{ else }
	\end{array}
	\right.
\)
be the \textbf{characteristic function} interpreting 
$\event_{(s,t,A)(s,r,B)(r+1,t,C)}$ as 
a simple subset of the set of all possible syntax 
trees $\trees(y)$ of the sentence $y$.
Thus, the frequency $f_{A \rightarrow B C}(x)$ 
of the rule $A \rightarrow B\ C$ occurring in the syntax tree 
$x \in \trees(y)$ can be computed as follows:
\[
	f_{A \rightarrow B C} (x)
	\ = \
	\sum_{1\le s \le r < t \le n}
	f_{(s,t,A)(s,r,B)(r+1,t,C)}(x)
	\ .
\]
Using the \textbf{linear properties of the expected frequencies}
\(
	\expect{p(.|y)}{.}	
	,
\)
it follows:
\begin{eqnarray*}
	\expect{\dist(.|y)}
	{
		f_{A \rightarrow B C}
	}
	& = &
	\expect{\dist(.|y)}
	{
		\sum_{1\le s \le r < t \le n}
		f_{(s,t,A)(s,r,B)(r+1,t,C)}
	}
\\
	& = &
	\sum_{1\le s \le r < t \le n}\
	\expect{\dist(.|y)}
	{
		f_{(s,t,A)(s,r,B)(r+1,t,C)}
	}
\\
	& = &
	\sum_{1\le s \le r < t \le n}\
	\sum_{x \in \trees(y)}\
	\dist(x|y) 
	\cdot 
	f_{(s,t,A)(s,r,B)(r+1,t,C)}(x)
\\
	& = &
	\frac{1}{p(y)}
	\sum_{1\le s \le r < t \le n}\
	\sum_{x \in \trees(y)}\
	\dist(x) 
	\cdot 
	f_{(s,t,A)(s,r,B)(r+1,t,C)}(x)
\\
	& = &
	\frac{1}{p(y)}
	\sum_{1\le s \le r < t \le n}\
	\sum_{x \in \event_{(s,t,A)(s,r,B)(r+1,t,C)}}\
	\dist(x)
\\
	& = &
	\frac{1}{p(y)}
	\sum_{1\le s \le r < t \le n}\
	\dist(\event_{(s,t,A)(s,r,B)(r+1,t,C)}) 
\\
	& = &
	\frac{1}{P}
	\sum_{1\le s \le r < t \le n}\
	f(s,t,A) \cdot
	\dist(A \rightarrow B C) \cdot
	e(s,r,B) \cdot
	e(r+1,t,C)
\\
	& = &
	C_y(A \rightarrow B\ C)
	\ .
\end{eqnarray*} 
The second case, for rules of the form 
$A \rightarrow a$, 
follows analogously with spans $(s,s,A)$ and
$(s,s,a)$. Here, the details are omitted, but see 
\cite{Prescher:DynamicEM} \qed


\end{document}